\documentclass[10pt,twocolumn,letterpaper]{article}

\usepackage{cvpr}
\usepackage{times}
\usepackage{epsfig}
\usepackage{graphicx}
\usepackage{amsmath}
\usepackage{amssymb}
\usepackage{textcomp}
\usepackage{gensymb}
\usepackage[linesnumbered,algoruled,boxed,lined]{algorithm2e}
\usepackage{authblk}

\cvprfinalcopy 

\def\cvprPaperID{****} 

\newcommand{\norm}[1]{\left\lVert#1\right\rVert}
\begin{document}

\title{Low-Cost Transfer Learning of Face Tasks}

\author[1]{Thrupthi Ann John}
\author[1]{Isha Dua}
\author[2]{Vineeth N Balasubramanian}
\author[1]{C. V. Jawahar}
\affil[1]{IIIT Hyderabad}
\affil[2]{IIT Hyderabad}


\maketitle

\begin{abstract}
   Do we know what the different filters of a face network represent? Can we use this filter information to train other tasks without transfer learning? For instance, can age, head pose, emotion and other face related tasks be learned from face recognition network without transfer learning? Understanding the role of these filters allows us to transfer knowledge across tasks and take advantage of large data sets in related tasks. Given a pretrained network, we can infer which tasks the network generalizes for and the best way to transfer the information to a new task.

We demonstrate a computationally inexpensive algorithm to reuse the filters of a face network for a task it was not trained for. Our analysis proves these attributes can be extracted with an accuracy comparable to what is obtained with transfer learning, but 10 times faster. We show that the information about other tasks is present in relatively small number of filters.  We use these insights to do task specific pruning of a pretrained network.  Our method gives significant compression ratios with reduction in size of 95\% and computational reduction of 60\% 
\end{abstract}

\section{Introduction}
   
Deep neural networks are very popular in machine learning, achieving state-of-the-art results in most modern machine learning tasks. A key reason for their success has been attributed to their capabilities to learn appropriate feature representations for a given task. The features of deep networks have also been shown to generalize across various tasks \cite{MidlevelTransfer,general1}, and learn information about tasks which they did not encounter during training. This is possible because tasks are often related, and when a deep neural network learns to predict a given task, the feature representations it learns can be adapted to other similar tasks to varying degrees. Several efforts in recent years  ~\cite{intro_aux4, intro_aux5, intro_aux2, intro_aux3} have found such relationships between tasks that are diverse but related, such as object detection to image correspondence \cite{intro_aux2}, scene detection to object detection \cite{intro_aux3} and expression recognition to facial action units \cite{intro_aux5}. 

\begin{figure}
    \centering
    \includegraphics[width=\linewidth]{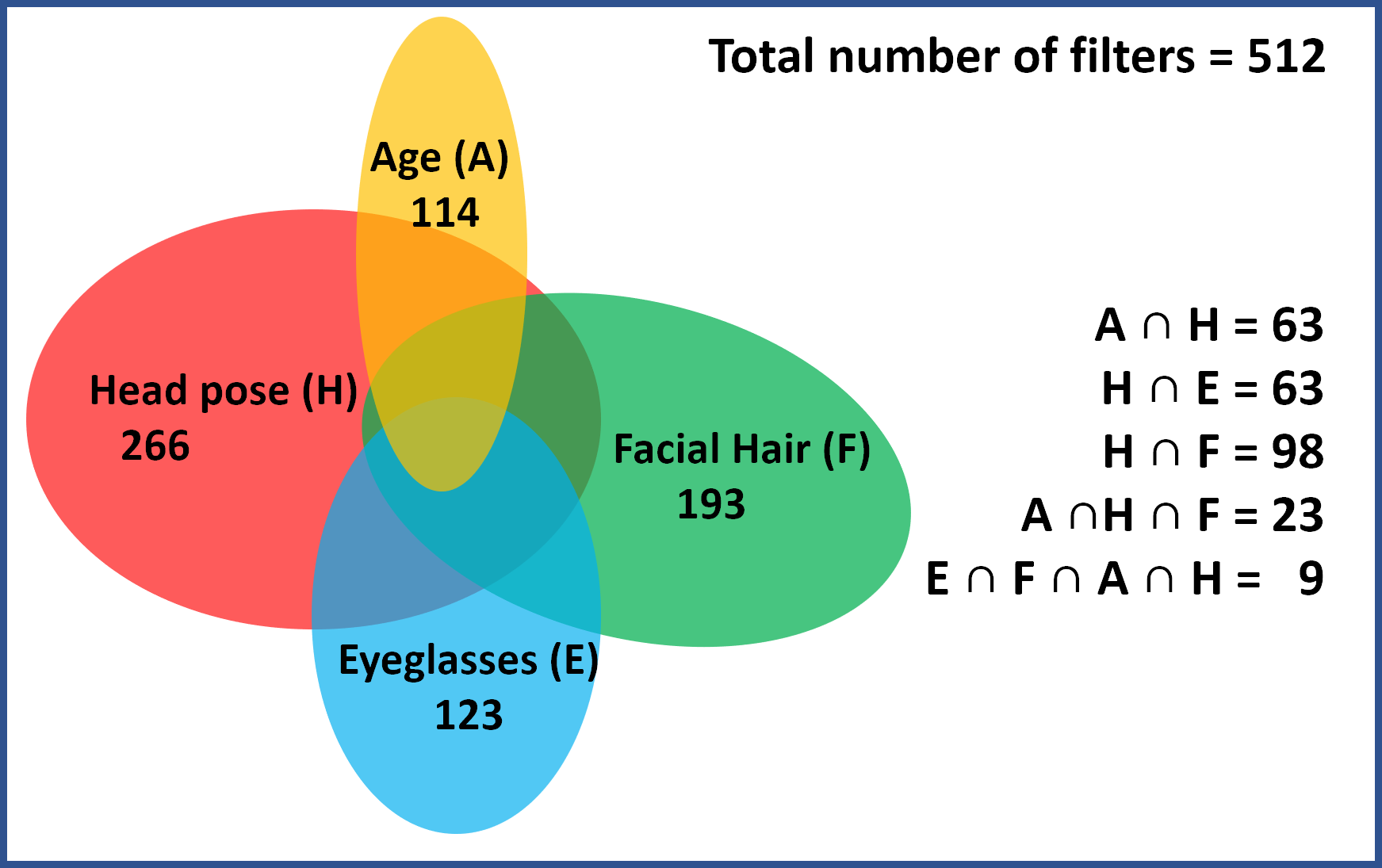}
    \caption{Figure depicts how information about different face tasks overlaps in the last convolutional layer of a face recognition network. The outer rectangle represents all the filters of the last layer, whereas the ovals depict the filters which contain information about different tasks. We observe that most of the tasks are encoded using very few filters, thus allowing us to compress the network by removing redundant filters.}
    \label{fig:venndiagram}
\end{figure}

One of the most important domains in computer vision is the face domain. Tasks in the face domain such as face recognition and emotion detection are very important in many applications such as biometric verification, surveillance and human-computer interaction. These tasks tend to be quite challenging, as face images can be similar to each other and face tasks often involve fine-grained classification. Besides,for a given face task, many variations need to be taken into account. For example, recognition has to be invariant to changes in expression, pose and accessories worn around the face. In recent years, we have seen that deep networks handle these challenging tasks very well.  Deep Face Recognition \cite{VGGFace} which trains a  VGG-16\cite{VGG16} model on 2.6 million face images gives an accuracy of 97.27 on the LFW data set\cite{lfw}, whereas FaceNet\cite{facenet} which uses a pre-trained deep convolutional network to learn the embedding instead of an intermediate bottleneck layer gives an accuracy of 99.67\%. 

It is no new fact that tasks in the face domain are highly related to each other. As much as face tasks have to deal with many variations in images, from another perspective, different face tasks (such as face recognition, pose estimation, age estimation, emotion detection) operate on input data that are fairly similar to each other. These face tasks attempt to capture fine-grained differences between the images.  Since the tasks are related and come from the same domain, one would expect that learning one task can help learn other tasks. We provide a simple generalizable framework to find  relationships between face tasks, which can help models trained on a face task be transferred with very few computations to other face tasks.

One approach to achieve the aforementioned objective that has been studied in literature is multitask learning \textsc{mtl} \cite{heteroface, hyperface,multitaskCNN, FLD}, where a single deep neural network is trained to solve multiple tasks given a single face image. An \textsc{mtl} architecture generally consists of a convolutional network which branches into multiple arms, each addressing a different face task. For example, Zhang \etal \cite{FLD} estimate  facial landmarks, head  yaw,  gender, smile and eyeglasses in a single deep model. These networks contain a large number of parameters and can be unwieldy to train. Furthermore, these methods require large data sets with labels for each of the considered tasks. In contrast, our studies on understanding what face networks learn and how they correlate with other face tasks, directly lend themselves to a method which can solve multiple tasks with far less labeled data and training overhead. 

Our framework is pivoted on understanding the role and information contained in different filters in a convolutional layer with respect to other tasks that the base network was not originally trained for. Consider the last convolutional layer of the trained VGG-Face \cite{VGGFace} model which has 512 convolutional filters. Different filters contain information about different face tasks. Figure \ref{fig:venndiagram} shows the distribution of these tasks in the 512 filters. We observe that while many filters are not relevant for other tasks, some filters are fairly general and can be easily adapted to solve other face tasks. Complementarily, we observe that when finetuning a pretrained network for a different face task, the task-specific filters (that do not show relevance for use in the other task) can be removed, resulting in a highly streamlined network without much reduction in performance. We provide a pruning algorithm that removes these redundant filters so that the network can be used for a task it was not originally trained for. We achieve up to 98\% reduction in size and 78\% reduction in computational complexity with comparable performance. 


Our key contributions are summarized as follows:
\begin{enumerate}
    \item We introduce a simple method to analyze the internal representation of a pretrained face network and how information about other related tasks it was not trained for, is encoded in this network.
    \item We present a computationally inexpensive method to transfer the information to other tasks without explicit transfer learning.
    \item We show that information about other tasks is concentrated in very few filters. This knowledge can be leveraged to achieve cross-task pruning of pre-trained networks, which provides significant reduction of space and computational complexity.
\end{enumerate}

\section{Related Work}

Deep neural networks achieve state-of-the-art results in many areas, but are notoriously hard to understand and interpret. There have been many attempts to shed light on the internal working of deep networks and the semantics of the features it generates. One class of methods \cite{visual7, visual4, visual10, visual18,visual9, visual11, visual8} visualizes the convolutional filters of CNNs by mapping them to the image domain. Other works \cite{gradcam2, gradcam1, gradcam6, gradcam5, gradcam4, gradcam3} attempt to map parts of the image a network pays attention to, using saliency maps. 

Other efforts attempt to interpret how individual neurons or groups of neurons work. Notably, the work by Raghu \etal \cite{neuraldeletion2} proposed a method to determine the true dimensionality of a layer and determined that it is much smaller than the number of neurons in that layer. S. Morcos \etal \cite{neuraldeletion1} studied the effect of single neurons on generalization performance by removing neurons from a neural network one by one. Alian and Bengio \cite{UnderstandingIL} developed intuition about the trained model by using linear classifiers that use the hidden units of a given intermediate layer as discriminating features. This allows the user to visualize the state of the model at multiple steps of training. D.Bau \etal \cite{netdissect2017} proposed a method for quantifying interpretability of latent representations of CNNs by evaluating the alignment between individual hidden units and a set of semantic concepts.These methods focus on interpreting the latent features of deep networks trained on a single task. In contrast, our work focuses on analyzing how the latent features contain information about external tasks which the network did not encounter during training, and how to reuse these features for these tasks.
 


We explore a simple alternative to transfer learning \cite{ transfer3, transfer4, transfer2} in this work. In traditional transfer learning, a network is trained for a base task, its features are transferred to a second network to be trained on a target data set/task. There are several successful examples of transfer learning in computer vision including \cite{transfer5, transfer6, MidlevelTransfer, transfer7}. Zamir \etal \cite{taskonomy} used a computational approach to recommend the best transfer learning policy between a set of source and target tasks. They also find structural relationships between vision tasks using this approach. Yosinski \etal \cite{HowTransferable} provided many recommendations for best practices in transfer learning. They quantified the degree to which a particular layer is general or specific, i.e., how well features at that layer transfer from one task to another. They also quantified the `distance' between different tasks using a computational approach. In contrast, we provide a computationally cheaper alternative that emerges from understanding the filters of a convolutional network. We now present our motivation and methodology.


\section{Learning Relationships between Face Tasks}
\label{sec:regression}
It has been widely known in computer vision that CNNs learn generic features that can be used for various related tasks. For example, when a CNN learns to recognize faces, the convolutional filters may automatically learn to predict various other facial attributes such as head pose, age, gender etc. as a consequence of learning the face recognition task. These `shared' filters can be reused for tasks other than what the network was trained for. This is our key idea.

\begin{figure}
    \centering
    \includegraphics[width=\linewidth]{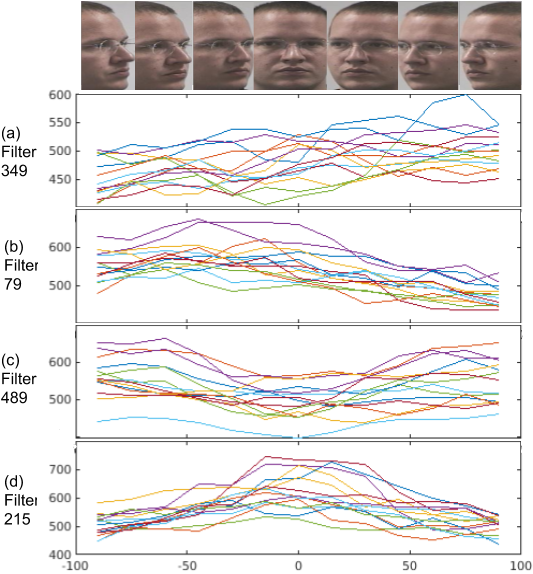}
    \caption{Figure shows the correlation between yaw angle on Head Pose Image Database and average responses of a few convolutional filters from the last layer of VGG-Face. The different lines in each graph represent 15 different identities: (a) high activation for left-facing faces; (b) high response for faces facing right; (c) high response for sideways faces; (d) high response for frontal faces}
    \label{fig:correlation}
\end{figure}
We study the generalizability of such features using the following experiment. Consider the VGG-Face network \cite{VGGFace} trained for face recognition on 2.6 million images. We examine how the features of this network generalize to the task of determining head pose. The Head Pose Image Database \cite{headpose} is a benchmark of 2790 monocular face images of 15 persons with pan angles from $-90\degree$ to $+90\degree$ at $15\degree$ intervals and various tilt angles. We use this data set because all the attributes are kept constant except for the head pose. We pass the images of the data set through the VGG-Face network and study the $L_2$-norm of feature maps of each of the 512 filters in the last convolutional layer. We observe in Figure \ref{fig:correlation} that the response of certain filters are correlated to the yaw of the head. Some filters give high response for faces looking straight, whereas other filters give high response for left-facing faces. This experiment shows that few of the filters of a network trained for face recognition encode information about yaw without being explicitly trained for yaw angle estimation during training. This motivates the need for a methodology to identify and exploit the use of such filters to predict related tasks on which the original network was not trained.
\begin{table}[!h]
    \centering
    \begin{tabular}{|p{1.5cm}|p{4cm}|p{1.5cm}|}
    \hline \hline
       Task  & Type & Label Source \\
       \hline \hline
        Identity & Categorical (10177 classes) & CelebA \cite{CelebA} \\
        \hline
        Gender & Categorical (2 classes) & CelebA\cite{CelebA} \\
        \hline
        Facial Hair& Multilabel (5 classes: 5 o'clock shadow, goatee,
        sideburns, moustache, no beard) & CelebA \cite{CelebA}\\
        \hline
      Accessories 
       & Multilabel (5 classes: earrings, hat, necklace, necktie, eyeglasses) & CelebA \cite{CelebA} \\ \hline
        Age & Categorical (10 classes)  & Imdb-Wiki \cite{Age_CelebA}\\ \hline
        Emotions & Categorical (7 classes: angry, disgust, fear,happy, sad, surprise,neutral) &  Fer13 \cite{Emotion_CelebA}\\ \hline
        Head pose & Categorical (9 classes) & 3DMM\cite{Headpose_CelebA} \\
         \hline \hline
    \end{tabular}
     \caption{List of different tasks and corresponding labels. The labels for the first four tasks are provided with the CelebA data set. The other labels were obtained using known methods \cite{Headpose_CelebA, Emotion_CelebA, Age_CelebA} }
    \label{tab:dataset}
\end{table}
\paragraph{Methodology:} Let a dictionary of face tasks be defined by $F = \{f_1, f_2, \ldots, f_n\}$. Let $f'$ be the primary task. Then the set of satellite tasks are denoted by $F - \{f'\}$. We train a network model $\mathcal{M}$ on $f'$ and use its features to regress for a satellite task $f^t \in F - \{f'\}$. For example, we can train a network on the primary task of face recognition and use its features to regress for satellite tasks such as age, head pose and emotion detection. To this end, we consider a convolutional layer of model $\mathcal{M}$ with, say, $k$ filters. Let the activation map of layer $l$ on image $I$ be denoted by $A^l(I)$, and has size $k \times u \times v$, where each activation map is of size $u \times v$. We hypothesize that, unlike contemporary transfer learning methodologies (that finetune the weights or input these activation maps through further layers of a different network), a simple linear regression model is sufficient to obtain the predicted label of the satellite task, $f^t$. Our procedure is outlined in Algorithm \ref{alg:regress}. First, we take the activation map of a convolutional layer and perform global average pooling on it. This is then used as a feature vector to regress the satellite tasks. Typically, a large data set is used to train the primary task, as with any other deep face network. However, owing to the simplicity of our satellite task model, limited data is sufficient to train the satellite model using linear regression.
\begin{algorithm}
	\DontPrintSemicolon
	\KwIn{\\
	Face image data set for satellite task $f^t \in F - \{f'\}$: $\{I_1, \ldots, I_n\}$ with corresponding ground truth $Y^t=\{y^t_1, \ldots, y^t_n\}$\\ 
	$\mathcal{M}$ is a model obtained by training for the primary face task, $f'$. \\
	$A^l(I_j)$ is the activation map (size $u \times v$) of layer $l$ with $k$ filters on image $I_j$, $j=1,\cdots,n$\\
	}
	\KwOut{Regression model $W^t$ for $f^t$}
	$W^t = \underset{W}{\arg \min}\sum_{j=1}^n \frac{1}{2} \norm{w^T A^l(I_j) - y^t_j}_2^2$
	
	\caption{Training Satellite Face Task Model from Primary Task Model}
	\label{alg:regress}
\end{algorithm}

  \begin{figure}
    \centering
    \includegraphics[width=0.9\linewidth ]{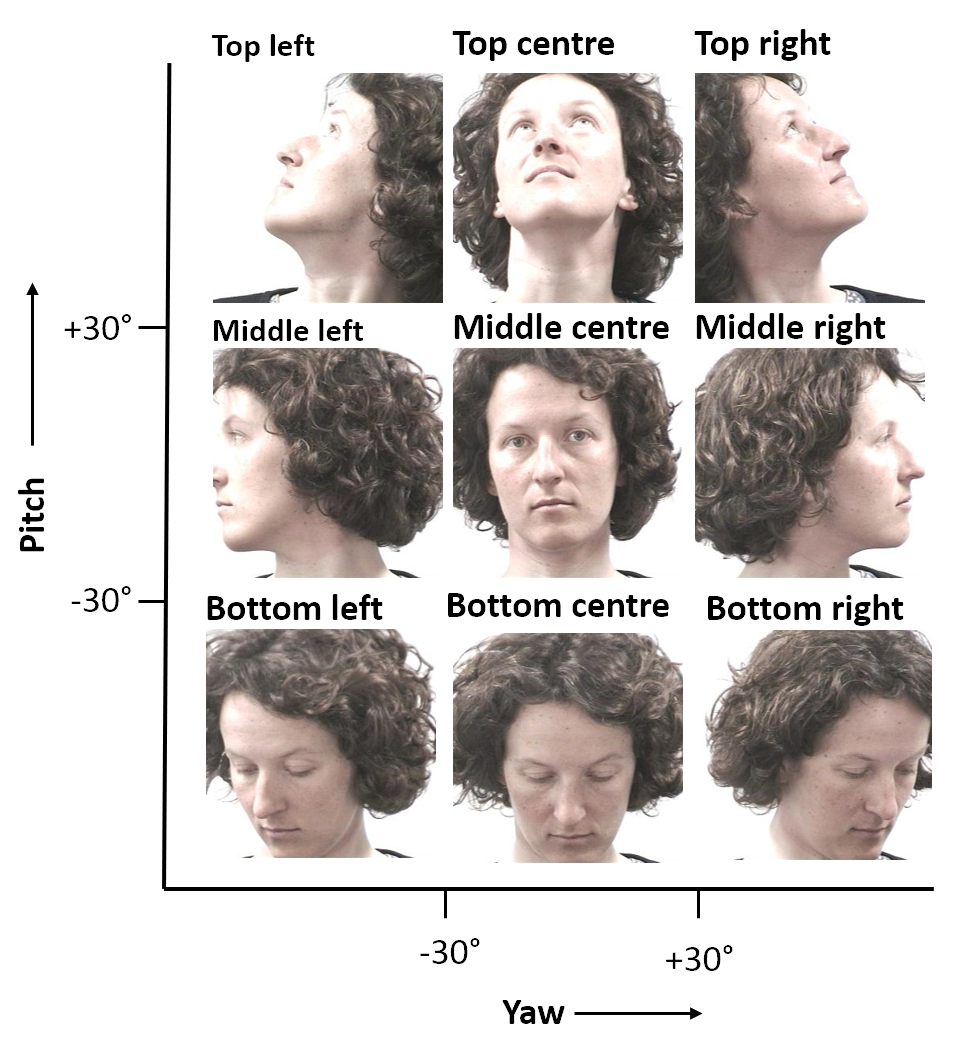}
    \caption{Classes for head pose task in Section \ref{sec:regression}. The yaw and pitch were divided into bins with 60\degree{}  bin size.}
    \label{fig:headposelabel}
\end{figure}
To validate the above mentioned method, a data set with ground truth for all considered face tasks is essential. We used the CelebA data set \cite{CelebA}, which consists of 202,599 images all experiments in Section \ref{sec:regression}.  The labels for identity, gender, accessories and facial hair are available as a part of the data set. For age, emotion and pose, we generated the ground truth using known methods. The ground truth for age was obtained using the method DEX: Deep EXpectation of apparent age from a single image \cite{Age_CelebA}. This method uses a VGG16 architecture and was trained on the IMDB-WIKI data set which consists of 0.5 million images of celebrities crawled from IMDB and Wikipedia. The ages obtained using this method were binned into 10 bins, with each bin having 10 ages. Head pose was obtained by registering the face to a 3D face model using linear pose fitting \cite{Headpose_CelebA}. The model used is a low-resolution shape-only version of the Surrey Morphable Face Model. The yaw and pitch values were binned into 9 bins ranging from top-left to bottom-right. The binned pose values are depicted in Figure \ref{fig:headposelabel}. For emotion, a VGG-16 model was trained on FER 2013 data set \cite{Emotion_CelebA} with 7 classes. (See Table \ref{tab:dataset} for the details of the considered data set). 

\begin{figure}
    \centering
     \includegraphics[width = \linewidth]{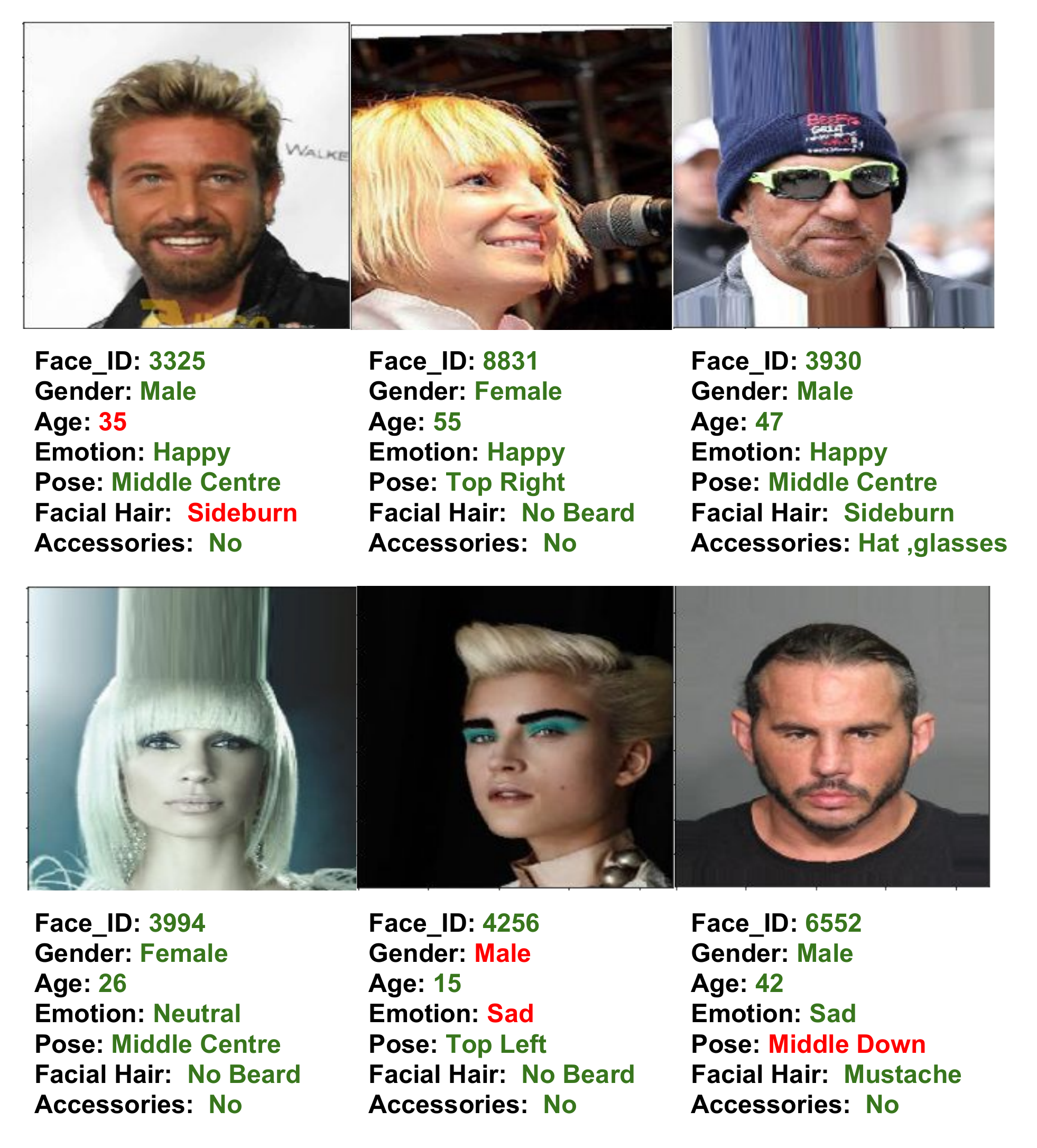}
    \caption{Sample results obtained on CelebA data set using linear regression on the activation maps of a CNN trained for face recognition. Green text shows correct predictions, and red text shows incorrect predictions.}
    \label{fig:qualitative1}
\end{figure}

\begin{figure*}
    \centering
     \includegraphics[width = 0.95\linewidth]{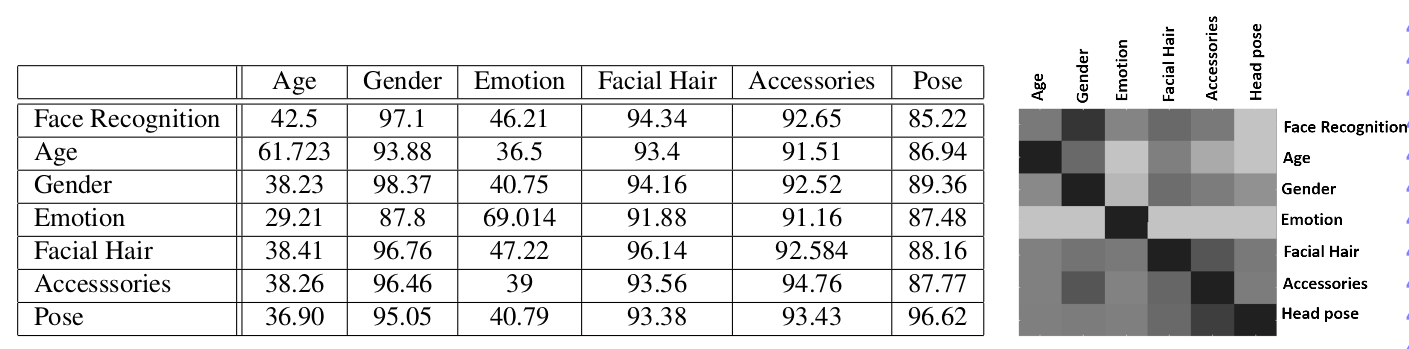}
    \caption{Table (left) shows results of transferring tasks using the proposed method. Each rows corresponds to the primary task and the columns correspond to the satellite task. We report the accuracy obtained when transferring a network pre-trained on the primary task to the considered satellite task. The diagonal cells show the accuracy obtained  while training the primary task. The figure (right) shows a heatmap of transfer capability of one face task to another based on this methodology (darker is better; for example, face recognition models can regress gender very well, while age estimation models are one of the least capable of estimating emotions).}
    \label{tbl:regression}
\end{figure*}

We consider the seven face tasks listed in Table \ref{tab:dataset}. The entire CelebA data set was divided into 50\% train, 25\% validation and 25\% test sets. We used a pre-trained VGG Face model \cite{VGGFace} and finetuned it for a considered primary task. The ground truth for the satellite tasks were created by taking a subset consisting of 20,000 images from CelebA ($\approx 10\%$ of the data set). This was also divided into 50\% train, 25\% validation and 25\% test sets. All our reported results are obtained by averaging over three random trials, obtained by different partitions of the satellite data. 
We converted the continuous regression outputs into categorical attributes for each of the tasks. For binary classes such as gender, our output was regressed to a value between 0 and 1, and a threshold (learned on the validation set) was used to decide the label on the test set. For multicategory classes such as age and pose, we regressed to a continuous label space based on the original labels. We then binned it using the same criteria we used for training the primary networks.

In order to determine how well a particular transfer took place, we compared the performance of our models learned on satellite tasks to the accuracy obtained by a network which was trained explicitly on the same task as primary. For example, say we want to compare the accuracy obtained by transferring a network trained for face recognition to the gender task. We do this by comparing the regression accuracy of face recognition$\rightarrow$gender with the network trained on the full data set for gender. This is measured as percentage reduction in performance when changing from the full data set to the subset. 

Figure \ref{tbl:regression}(left) shows the results of transferring tasks using regression. The activations were regressed to continuous labels, which were then binned to get the accuracy.  For the emotion detection task, we used linear classification. The primary tasks are represented by each row, which was then transferred to each of the satellite tasks represented by the columns. The accuracy obtained by a network trained for the primary task is denoted in the cells where the primary and secondary task are the same. For each satellite task, percentage reduction in performance compared to the network trained on the corresponding primary task is also captured in Figure \ref{tbl:regression} (right), with lower values (darker cells) being better. We show some qualitative results obtained using our regression algorithm in Figure \ref{fig:qualitative1}. 

We notice that networks trained on primary tasks give better results while regressing with tasks with which the primary tasks may have some correlation. For example, a model trained on gender recognition as the primary task gives good results for facial hair estimation and vice versa (supports common knowledge). 
Similarly, the accessories and gender estimation tasks are strongly correlated because certain accessories such as neck tie, earrings, and necklace are correlated strongly with gender. On the other hand, emotion gives low accuracy for all other tasks, since emotion is usually learned independently from other facial attributes.  Face recognition gives very good results for gender, facial hair and accessories since these vary from individual to individual. Face recognition does not give the best results for pose, because face recognition has to be invariant to pose. Curiously, age is regressed well by the face recognition network. This may be due to biases in the data set, where images belonging to  each individual do not have a large range of ages.

\begin{figure}
    \centering
     \includegraphics[width = 1.04\linewidth]{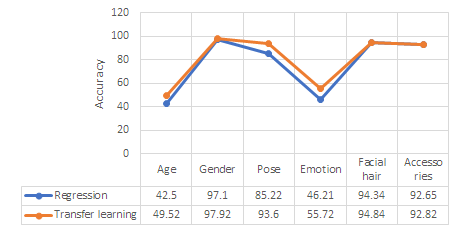}
      \caption{Graph shows comparison between regression and transfer learning for a network pretrained on face recognition and transferred to six other tasks. We can see that results for regression and transfer learning are very close, thereby allowing us to effectively replace transfer learning with our  method.}
    \label{fig:transfer}
\end{figure}

\paragraph{Relation to Transfer Learning:} We conducted experiments to examine how well our regression method compares to using transfer learning on various tasks. For this setting, we used networks pretrained on the face recognition tasks and reinitialized all the fully connected layers. We then froze the convolutional layers and trained the linear layers for the satellite tasks. The results can be seen in figure \ref{fig:transfer}. We can see that the regression results are close to the transfer results. Thus, we can use our simple regression method to find task relationships instead of doing expensive transfer learning for each task. Our regression method takes 10 seconds to run for a single task, as opposed to transfer learning, which takes 780 seconds. We thus achieve a speed-up of \textbf{78X} using our method.

\section{Pruning across Face Tasks}
\label{sec:pruning}
\paragraph{Motivation:} We have seen how the filters of convolutional layer of a pretrained network trained for one task can be repurposed for another. All filters of a layer may not have equal importance in terms of usefulness for predicting the satellite task.  In order to discover which filters from the  layer are useful for the task and which filters are redundant, we need to rank every group of filters  according to the accuracy obtained on regressing to the satellite task. Instead of exhaustively checking all groups of filters in a layer, we use feature selection to achieve this. In particular, we use \textsc{lasso} \cite{tibshirani1996regression}, a $L_1$-regularized regression method which selects only a subset of the variables in the final model rather than using all of the variables using the objective:
\begin{figure}
    \centering
    \includegraphics[width=\linewidth]{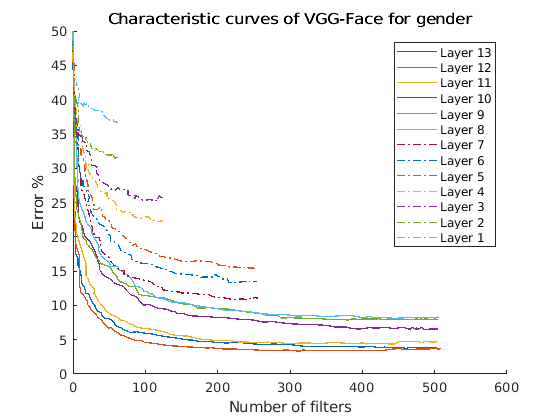}
    \caption{Characteristic curves for VGG-Face pretrained network regressed on gender. We can observe that regression gives very low error using as few as $\sim$100 filters. Adding more filters to the regression model does not have a large impact on the error, indicating that the additional filters do not capture much information about gender}
    \label{fig:characteristiccurve}
\end{figure}
	\begin{equation}
	\min_{\beta_0, \beta}\bigg(\frac{1}{2N}\sum_{i=1}^{N}(y_i - \beta_0 - x^T_i\beta^T_i)^2 + \lambda\sum_{j=1}^{p}|\beta_i|\bigg) 
	\label{eqn:lasso}
    \end{equation}
	where N is the number of observations, $y_i$ is the dependent variable  at observation i, $x_i$ is the independent variable (a vector of globally averaged filter responses at observation i) and $\lambda$ is a non-negative regularization parameter which determines the sparseness of the  regression weights $\beta$.
    
    As $\lambda$ increases, the number of filters chosen decreases, which are the non-zero coefficients of $\beta$. We train lasso using 100 different values of $\lambda$ to get a characteristic curve. The largest value of $\lambda$ is one that just makes all coefficients zero. The rest of the $\lambda$ values are chosen using a geometric sequence such that the ratio of largest to smallest $\lambda$ is 1e4.  For each layer, we get 100 regression models, each using different number of filters. The root mean squared error of the regression models is plotted with respect to the number of filters to obtain the characteristic curve of the layer. 
 
The characteristic curves of a network with respect to a satellite task T tell us how the information about T is distributed in the network. Let us observe the characteristic curves for VGG-Face pretrained network regressing on gender in Figure \ref{fig:characteristiccurve}. We can see that the error drops significantly using just a few filters and remains constant after that. This indicates that most of the information about gender is present in a few filters and the other filters are not needed for this task. We can use this fact to do cross-task pruning of the network by removing redundant filters.

 \begin{figure*}
    \centering
    \includegraphics[width=0.8\linewidth]{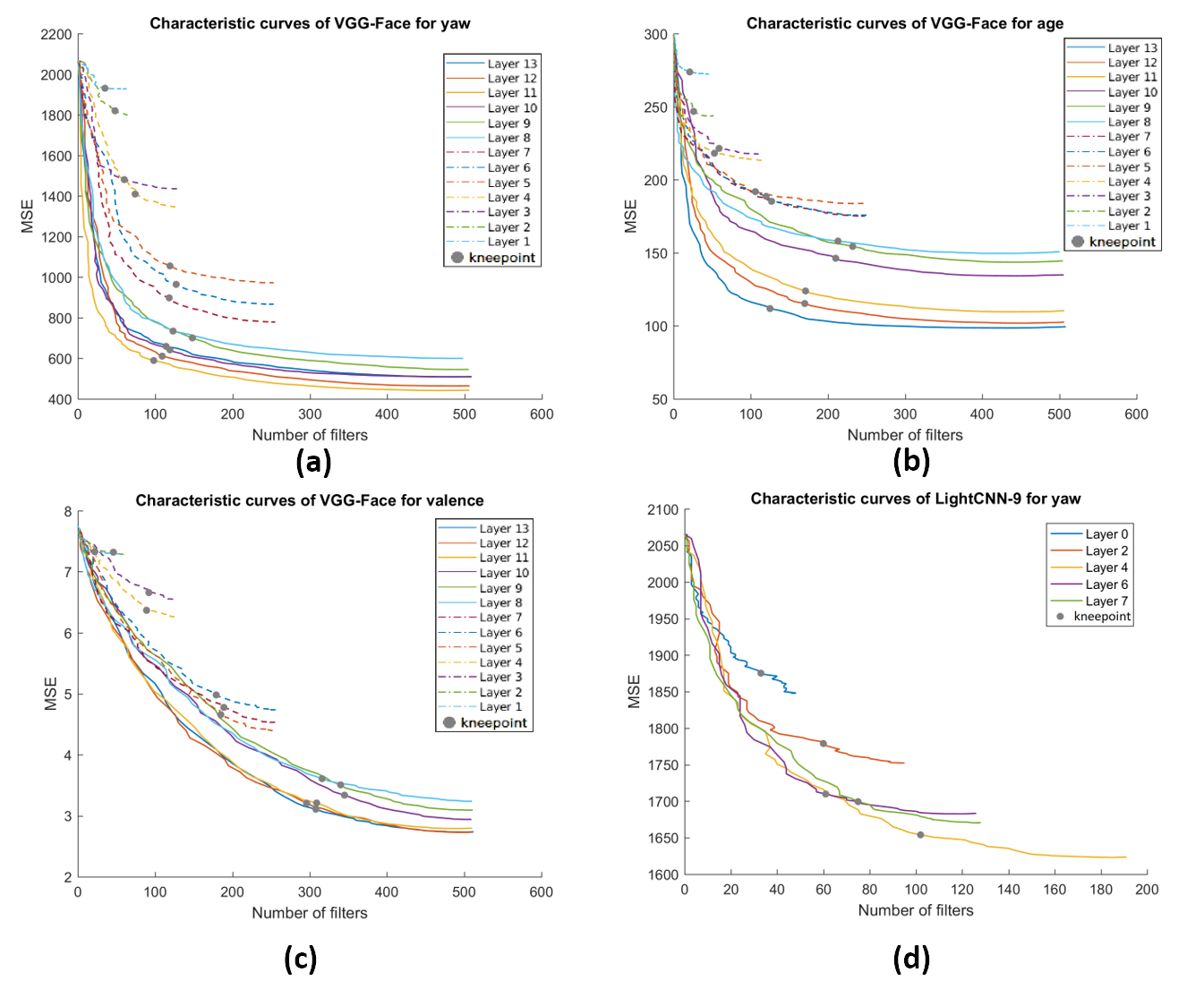}
    \caption{Characteristic curves obtained while regressing the primary network for various satellite tasks. The kneepoints indicate the regression model corresponding to threshold = 0.01. a.) VGG-Face regressed for head pose using AFLW data set \cite{AFLW} b.) VGG-Face regressed for age using AgeDB \cite{AgeDB} c.) VGG-Face regressed for valence (emotion) using AFEW-VA data set \cite{AFEWVA1,AFEWVA2} d.) LightCNN face network \cite{lightcnn} regressed for head pose using AFLW data set.(Zoom in to see the details)}
    \label{fig:characteristiccurves}
\end{figure*}

More examples of characteristic curves are given in figure \ref{fig:characteristiccurves}. Figure \ref{fig:characteristiccurves}A shows the characteristic curves obtained for VGG-Face pretrained network for the yaw task. These curves are quite sharp in the beginning, indicating that the yaw information is encoded by a few neurons. When we compare these to the characteristic curves of valence for VGG-Face network in Figure \ref{fig:characteristiccurves}C, we notice that these curves are very smooth and there is no elbow, showing that the information about valence is distributed throughout the layers. This is reflected by the compression ratios obtained while pruning for these tasks. (Refer to Table \ref{tbl:pruning}).

\paragraph{Pruning from Filters:}
The main goal of pruning is to reduce the size and computational complexity without much reduction in performance. Our pruning algorithm has two steps: 1. Remove top layers of the network which give less performance 2. For each layer, retain only filters that have information about task T. We use the characteristic curves as a guide for choosing which filters to keep and which to prune. We choose a knee-point for the characteristic curve of a layer in order to balance the number of filters and performance. The knee-point is defined as the minimum number of filters such that increase in RMSE is not more than a threshold. We have to minimize the number of features such that
\begin{equation}
    \text{RMSE}(k) - \min(m) < \gamma (\max(m) - \min(m))
    \label{eqn:kneepoint}
\end{equation}
where $m$ is the RMSE at a point on the curve, $k$ is the chosen knee-point and RMSE($k$) is the RMSE at the knee-point. $\gamma$ is a threshold expressed as a percentage of the range of RMSEs of a layer. We keep all the filters which are chosen by the regression model at $k$, and discard the rest using the procedure described in \cite{prune1}.

Let a convolutional layer $l$ have $o_l$ filters and $i_l$ input channels. The weight matrix of the layer will have the dimensions $o_l \times i_l \times k_l \times k_l$ (where $k_l \times k_l$ is the kernel size). Our procedure to prune filters from layer l is shown in Algorithm \ref{alg:prune}. 

\begin{table*}
		\begin{center}
			\begin{tabular}{|l|c|c|c|c|c|c|}
				\hline
				Attribute & Arch  & RMSE & FLOP & \% FLOP reduction & Parameters & \% Size reduction \\
				\hline\hline
                Yaw&VGG-Face&38.97&$2.54\times 10^{11}$&65.53&$2.03 \times 10^7$&96.50\\
                Age & VGG-Face & 14.28 & $3.28 \times 10^{11}$& 55.57 & $2.61 \times 10^7 $ &95.49 \\
                Valence & VGG-Face & 1.93 & $5.53 \times 10^{11}$ & 25.03 & $4.41 \times 10^7 $&92.39 \\
                Yaw & LightCNN & 40.42 & $7.09\times 10^{9}$ & 78.9 & $1.42 \times 10^6$ & 98.62\\
                
                \hline
                
			\end{tabular}
		\end{center}
		\caption{Table showing results of pruning, along with space and time compression ratios achieved by it. }
        \setlength{\belowcaptionskip}{-20pt}
		\label{tbl:pruning}
    \vspace{-10pt}
	\end{table*}

\begin{algorithm}
	\DontPrintSemicolon
   
	\KwIn{layer $l$, next convolutional layer $l+1$, $kn$: knee-point of the characteristic curve of layer $l$(Equation \ref{eqn:kneepoint})}
	\KwOut{Network with layers $l$ and $l+1$ pruned}
	$W_l \gets$ weights corresponding to the regression model at the knee-point. \;
	$n_l \gets$ number of non-zero elements of $W_l$  \;
	Let convolutional layer $l$ have $o_l$ filters and $i_l$ filters of size $k_l\times k_l$. Create a new convolution layer with $n_l$ filters. Its weight matrix is of size $n_l \times i_l \times k_l \times k_l$ \;
	 \For{ each non-zero element i in $W_l$}
     { \For{$j$ = 1:$n_l$, $p$ = 1:$i_l$, $q$ = 1:$k_l$, $r$ = 1:$k_l$}{$\text{newweights}[j,p,q,r] \gets \text{oldweights}[i,p,q,r]$  }}
	Replace layer $l$ with new convolutional layer \;
	\If{conv layer l+1 exists}
	{
		Create a new convolution layer with $n_l$ input channels.\;
		 \For{ each non-zero element i in $W_l$}
         {\For{$p$ = 1:$o_{l+1}$, $j$ = 1:$n_l$, $q$ = 1:$k_l$, $r$ = 1:$k_l$ } {$\text{newweights}[p,j,q,r] \gets \text{oldweights}[p,i,q,r]$ } }
		Replace layer $l+1$ with the new layer \;
	}		
	\caption{Prune filters from a layer in a network}
	 \label{alg:prune}
\end{algorithm}

In order to extend our pruning algorithm to architectures that are very different to VGG architecture, we have to modify the filter pruning procedure. Here, we examine the procedure to prune LightCNN-9 architecture. LightCNN introduces an operation called Max-Feature-Map (MFM) operation. An MFM 2/1 layer which has $i_l$ input channels and $o_l$ output channels has two components: a convolutional layer which has $i_l$ input channels and $2o_l$ output channels, and the MFM operator which combines the output channels using max across channels so that the output of the entire layer has only $o_l$ channels. Let the output X of the convolutional layer have dimensions $2o_l \times h \times w$. The MFM operator is defined as:
\begin{equation}
\hat{x }^p_{i,j} = max(x^p_{i,j}, x^{p+o_l}_{i,j})
\end{equation}
where $\hat{x }^p_{i,j}$ is the (i,j)\textsuperscript{th} element of channel $k$ of the output. If we want to keep $D = \{f_1, f_2, ..., f_d\}$ channels out of $o_l$ channels, we need to keep both 
$D$ and $D+o_l$ output channels from the convolutional layer and corresponding input channels from the next layer.

\begin{table}[]
    \centering
    \footnotesize
    \begin{tabular}{|c|c|c|c|}
    \hline
    Type   & \multicolumn{3}{|c|}{Gender}    \\
      \hline
         & Filter size/stride,pad & Output size & \#param  \\
         \hline \hline
        Conv2D & $3\times 3$ / 1,1 & $224 \times 224 \times 58$ & 6.4K   \\
        RelU & - & $224 \times 224 \times 58$ & - \\
        \hline
        Conv2D & $3\times 3$ / 1,1 & $224 \times 224 \times 54$ & 112.9K \\
        ReLU & - &  $224 \times 224 \times 54 $ & - \\
        \hline
        MaxPool & $2 \times 2 /2,0 $& $112 \times 112 \times 54 $&  - \\
        \hline
        Conv2D & $3\times 3$ / 1,1 & $112 \times 112 \times 109 $ & 212.3K  \\
        ReLU & - &  $112 \times 112 \times 109 $ & -  \\
        \hline
        Conv2D & $3\times 3$ / 1,1  & $112 \times 112 \times 114 $ & 447.8K \\
        ReLU & - &  $112 \times 112 \times 114 $ & -\\
        \hline
        MaxPool & $2\times 2$ / 2,0 &  $56 \times 56 \times 114 $ & - \\
        \hline
        Conv2D & $3\times 3$ / 1,1 & $56 \times 56 \times 216 $ & 887.3K  \\
        ReLU & - &  $56 \times 56 \times 216 $ & -  \\
        \hline
        Conv2D & $3\times 3$ / 1,1  & $56 \times 56 \times 223 $ & 1734.9K \\
        ReLU & - &  $56 \times 56 \times 223 $ & -\\
        \hline
        Conv2D & $3\times 3$ / 1,1  & $56 \times 56 \times 232 $ & 1863.4K \\
        ReLU & - &  $56 \times 56 \times 232 $ & -\\
        \hline
        MaxPool & $2\times 2$ / 2,0 &   $28 \times 28 \times 232 $ & - \\
        \hline
         Conv2D & $3\times 3$ / 1,1 & $28 \times 28 \times 373 $ & 3116.8K  \\
        ReLU & - &  $28 \times 28 \times 373 $ & -  \\
        \hline
        Conv2D & $3\times 3$ / 1,1  & $28 \times 28 \times 373 $ & 5010.1K \\
        ReLU & - &  $28 \times 28 \times 373 $ & -\\
        \hline
        Conv2D & $3\times 3$ / 1,1  & $28 \times 28 \times 379 $ & 5090.7K \\
        ReLU & - &  $28 \times 28 \times 379 $ & -\\
        \hline
        MaxPool & $2\times 2$ / 2,0 &  $14 \times 14 \times 373 $ & - \\
        \hline
         Conv2D & $3\times 3$ / 1,1  & $14 \times 14 \times 302 $ & 4121.7K \\
        ReLU & - &  $14 \times 14 \times 302 $ & -\\
         \hline
         Conv2D & $3\times 3$ / 1,1  & $14 \times 14 \times 309 $ & 3360.6K \\
        \hline
         Avg. pool & - & 309 & - \\
         \hline
         Linear & - & 2 & 2.4K \\
         \hline
    \end{tabular}
    \caption{The Architecture of VGG-Face network after pruning for gender}
    \label{tbl:genderarch}
\end{table}

\begin{table}[]
    \centering
    \footnotesize
    \begin{tabular}{|c|c|c|c|}
    \hline
    Type   & \multicolumn{3}{|c|}{Age}    \\
      \hline
         & Filter size/stride,pad & Output size & \#param  \\
         \hline \hline
        Conv2D & $3\times 3$ / 1,1 & $224 \times 224 \times 60$ & 6.7K   \\
        RelU & - & $224 \times 224 \times 60$ & - \\
        \hline
        Conv2D & $3\times 3$ / 1,1 & $224 \times 224 \times 58$ & 125.5K \\
        ReLU & - &  $224 \times 224 \times 58 $ & - \\
        \hline
        MaxPool & $2 \times 2 /2,0 $& $112 \times 112 \times 58 $&  - \\
        \hline
        Conv2D & $3\times 3$ / 1,1 & $112 \times 112 \times 79 $ & 165.2K  \\
        ReLU & - &  $112 \times 112 \times 79 $ & -  \\
        \hline
        Conv2D & $3\times 3$ / 1,1  & $112 \times 112 \times 105 $ & 299K \\
        ReLU & - &  $112 \times 112 \times 105 $ & -\\
        \hline
        MaxPool & $2\times 2$ / 2,0 &  $56 \times 56 \times 105 $ & - \\
        \hline
        Conv2D & $3\times 3$ / 1,1 & $56 \times 56 \times 201 $ & 760.5K  \\
        ReLU & - &  $56 \times 56 \times 201 $ & -  \\
        \hline
        Conv2D & $3\times 3$ / 1,1  & $56 \times 56 \times 220 $ & 1592.8K \\
        ReLU & - &  $56 \times 56 \times 220 $ & -\\
        \hline
        Conv2D & $3\times 3$ / 1,1  & $56 \times 56 \times 240 $ & 1901.7K \\
        ReLU & - &  $56 \times 56 \times 240 $ & -\\
        \hline
        MaxPool & $2\times 2$ / 2,0 &   $28 \times 28 \times 240 $ & - \\
        \hline
         Conv2D & $3\times 3$ / 1,1 & $28 \times 28 \times 396 $ & 3423.0K  \\
        ReLU & - &  $28 \times 28 \times 396 $ & -  \\
        \hline
        Conv2D & $3\times 3$ / 1,1  & $28 \times 28 \times 449 $ & 6402.7K \\
        ReLU & - &  $28 \times 28 \times 449 $ & -\\
        \hline
        Conv2D & $3\times 3$ / 1,1  & $28 \times 28 \times 387 $ & 6257.0K \\
        ReLU & - &  $28 \times 28 \times 387 $ & -\\
        \hline
        MaxPool & $2\times 2$ / 2,0 &  $14 \times 14 \times 387 $ & - \\
        \hline
         Conv2D & $3\times 3$ / 1,1  & $14 \times 14 \times 332 $ & 4626.7K \\
        ReLU & - &  $14 \times 14 \times 332 $ & -\\
        \hline
         Conv2D & $3\times 3$ / 1,1  & $14 \times 14 \times 350 $ & 4184.6K \\
        ReLU & - &  $14 \times 14 \times 350 $ & -\\
        \hline
         Conv2D & $3\times 3$ / 1,1  & $14 \times 14 \times 275 $ & 3466.1K \\
        \hline
        Avg. pool & - & 275 & - \\
         \hline
         Linear & - & 10 & 11.04K \\
         \hline
    \end{tabular}
    \caption{The Architecture of VGG-Face network after pruning for age}
    \label{tbl:agearch}
\end{table}

\begin{table}[]
    \centering
    \footnotesize
    \begin{tabular}{|c|c|c|c|}
    \hline
    Type   & \multicolumn{3}{|c|}{Head pose}    \\
      \hline
         & Filter size/stride,pad & Output size & \#param  \\
         \hline \hline
        Conv2D & $3\times 3$ / 1,1 & $224 \times 224 \times 23$ & 2.5K   \\
        RelU & - & $224 \times 224 \times 23$ & - \\
        \hline
        Conv2D & $3\times 3$ / 1,1 & $224 \times 224 \times 12$ & 9.984K \\
        ReLU & - &  $224 \times 224 \times 12 $ & - \\
        \hline
        MaxPool & $2 \times 2 /2,0 $& $112 \times 112 \times 12 $&  - \\
        \hline
        Conv2D & $3\times 3$ / 1,1 & $112 \times 112 \times 121 $ & 52.756K  \\
        ReLU & - &  $112 \times 112 \times 121 $ & -  \\
        \hline
        Conv2D & $3\times 3$ / 1,1  & $112 \times 112 \times 110 $ & 479.6K \\
        ReLU & - &  $112 \times 112 \times 110 $ & -\\
        \hline
        MaxPool & $2\times 2$ / 2,0 &  $56 \times 56 \times 110 $ & - \\
        \hline
        Conv2D & $3\times 3$ / 1,1 & $56 \times 56 \times 234 $ & 927.5K  \\
        ReLU & - &  $56 \times 56 \times 234 $ & -  \\
        \hline
        Conv2D & $3\times 3$ / 1,1  & $56 \times 56 \times 230 $ & 1938.4K \\
        ReLU & - &  $56 \times 56 \times 230 $ & -\\
        \hline
        Conv2D & $3\times 3$ / 1,1  & $56 \times 56 \times 227 $ & 1880.4K \\
        ReLU & - &  $56 \times 56 \times 227 $ & -\\
        \hline
        MaxPool & $2\times 2$ / 2,0 &   $28 \times 28 \times 227 $ & - \\
        \hline
         Conv2D & $3\times 3$ / 1,1 & $28 \times 28 \times 370 $ & 3025.1K  \\
        ReLU & - &  $28 \times 28 \times 370 $ & -  \\
        \hline
        Conv2D & $3\times 3$ / 1,1  & $28 \times 28 \times 348 $ & 4636.7K \\
        ReLU & - &  $28 \times 28 \times 348 $ & -\\
        \hline
        Conv2D & $3\times 3$ / 1,1  & $28 \times 28 \times 390 $ & 4887.4K \\
        ReLU & - &  $28 \times 28 \times 390 $ & -\\
        \hline
        MaxPool & $2\times 2$ / 2,0 &  $14 \times 14 \times 390 $ & - \\
        \hline
         Conv2D & $3\times 3$ / 1,1  & $14 \times 14 \times 362 $ & 5083.9K \\
        ReLU & - &  $14 \times 14 \times 362 $ & -\\
        \hline
         Conv2D & $3\times 3$ / 1,1  & $14 \times 14 \times 395 $ & 5149.2K \\
        ReLU & - &  $14 \times 14 \times 395 $ & -\\
        \hline
         Conv2D & $3\times 3$ / 1,1  & $14 \times 14 \times 409 $ & 5817.6K \\
        \hline
        Avg. pool & - & 409 & - \\
         \hline
         Linear & - & 9 & 14.76K \\
         \hline
    \end{tabular}
    \caption{The Architecture of VGG-Face network after pruning for head pose}
    \label{tbl:headposearch}
\end{table}

\begin{table}[]
    \centering
    \footnotesize
    \begin{tabular}{|c|c|c|c|}
    \hline
    Type   & \multicolumn{3}{|c|}{Emotion}    \\
      \hline \hline
         & Filter size/stride,pad & Output size & \#param  \\
         \hline 
        Conv2D & $3\times 3$ / 1,1 & $224 \times 224 \times 53$ & 5.9K   \\
        RelU & - & $224 \times 224 \times 53$ & - \\
        \hline
        Conv2D & $3\times 3$ / 1,1 & $224 \times 224 \times 44$ & 84.128K \\
        ReLU & - &  $224 \times 224 \times 44 $ & - \\
        \hline
        MaxPool & $2 \times 2 /2,0 $& $112 \times 112 \times 44 $&  - \\
        \hline
        Conv2D & $3\times 3$ / 1,1 & $112 \times 112 \times 80 $ & 127.04K  \\
        ReLU & - &  $112 \times 112 \times 80 $ & -  \\
        \hline
        Conv2D & $3\times 3$ / 1,1  & $112 \times 112 \times 102 $ & 294.16K \\
        ReLU & - &  $112 \times 112 \times 102 $ & -\\
        \hline
        MaxPool & $2\times 2$ / 2,0 &  $56 \times 56 \times 102 $ & - \\
        \hline
        Conv2D & $3\times 3$ / 1,1 & $56 \times 56 \times 181 $ & 665.35K  \\
        ReLU & - &  $56 \times 56 \times 181 $ & -  \\
        \hline
        Conv2D & $3\times 3$ / 1,1  & $56 \times 56 \times 170 $ & 1108.4K \\
        ReLU & - &  $56 \times 56 \times 170 $ & -\\
        \hline
        Conv2D & $3\times 3$ / 1,1  & $56 \times 56 \times 203 $ & 1243.17K \\
        ReLU & - &  $56 \times 56 \times 203 $ & -\\
        \hline
        MaxPool & $2\times 2$ / 2,0 &   $28 \times 28 \times 203 $ & - \\
        \hline
         Conv2D & $3\times 3$ / 1,1 & $28 \times 28 \times 193 $ & 1411.2K  \\
        ReLU & - &  $28 \times 28 \times 193 $ & -  \\
        \hline
        Conv2D & $3\times 3$ / 1,1  & $28 \times 28 \times 286 $ & 1988.2K \\
        ReLU & - &  $28 \times 28 \times 286 $ & -\\
        \hline
        Conv2D & $3\times 3$ / 1,1  & $28 \times 28 \times 314 $ & 3234.2K \\
        ReLU & - &  $28 \times 28 \times 314 $ & -\\
        \hline
        MaxPool & $2\times 2$ / 2,0 &  $14 \times 14 \times 314 $ & - \\
        \hline
         Conv2D & $3\times 3$ / 1,1  & $14 \times 14 \times 218 $ & 2465.1K \\
        ReLU & - &  $14 \times 14 \times 218 $ & -\\
        \hline
         Conv2D & $3\times 3$ / 1,1  & $14 \times 14 \times 237 $ & 1860.9K \\
        ReLU & - &  $14 \times 14 \times 237 $ & -\\
        \hline
         Conv2D & $3\times 3$ / 1,1  & $14 \times 14 \times 225 $ & 1920.6K \\
        \hline
        Avg. pool & - & 225 & - \\
         \hline
         Linear & - & 7 & 6.3K \\
         \hline
    \end{tabular}
    \caption{The Architecture of VGG-Face network after pruning for emotion}
    \label{tbl:emotionarch}
\end{table}

LightCNN also has group layers which consist of two MFM layers. Consider a group layer with $i_l$ input channels, $o_l$ output channels and $k \times k$ filter size. The first MFM layer has $1\times 1$ convolutional layer with $i_l$ input layers and $i_l$ output layers. The second MFM layer has  $i_l$ input channels, $o_l$ output channels and $k \times k$ convolutional size. In order to keep $D$ filters in a group, we keep $D$ filters from the second MFM layer (as detailed above). We also keep $D$ filters from the first MFM layer of the next group.

\paragraph{Results:} Our pruning experiments are conducted on various data sets other than CelebA in order to determine if our insights can be adapted to other data sets. Our base network is the VGG-Face network \cite{VGGFace}, which we prune so that it can be reused for head pose, age and emotion (valence). We use Annotated Facial Landmarks in the Wild (AFLW) \cite{AFLW} data set for pose task. AFLW contains a large number of `in the wild' faces for which yaw, pitch and roll attributes are provided. We only use yaw values for our experiments.

For age prediction task, we use AgeDB data set \cite{AgeDB}, which has more than 15000 images with ages ranging from 1 to 101. The valence task is analyzed using AFEW-VA data set \cite{AFEWVA2, AFEWVA1}. This data set consists of clips extracted from feature films that have per frame annotation of valence and arousal. The data sets are randomly split into 75\% for training and 25\% for testing.
We perform the pruning experiments on VGG-Face \cite{VGGFace} for yaw, age and valence tasks. The results are given in Table \ref{tbl:pruning}. We can see that the computation and size of networks have been significantly reduced, while the error is less than the network that was trained from scratch. There are several reasons for this. First, the fully connected layers are removed, which reduces the number of parameters by a great amount. Next, as we can see from Figure \ref{fig:characteristiccurve}, the information pertaining to satellite tasks are concentrated in very few neurons in VGG-Face network. Hence we are able to remove many convolutional filters from each layer without impacting the performance much. For valence, the information is spread throughout the network, hence the compression ratio is less than that of yaw and age. The architectures of the pruned networks are given in Tables \ref{tbl:genderarch}, \ref{tbl:agearch}, \ref{tbl:headposearch} and \ref{tbl:emotionarch}.

Table \ref{tbl:pruning} also shows the results of  our pruning method on LightCNN architecture, in order to show that our algorithm is applicable for a variety of architectures. We pruned LightCNN network pretrained on face recognition for yaw tasks. As can be seen from Table \ref{tbl:pruning}, the algorithm works equally well for LightCNN architecture.

 \begin{figure*}
    \centering
    \includegraphics[width=\linewidth]{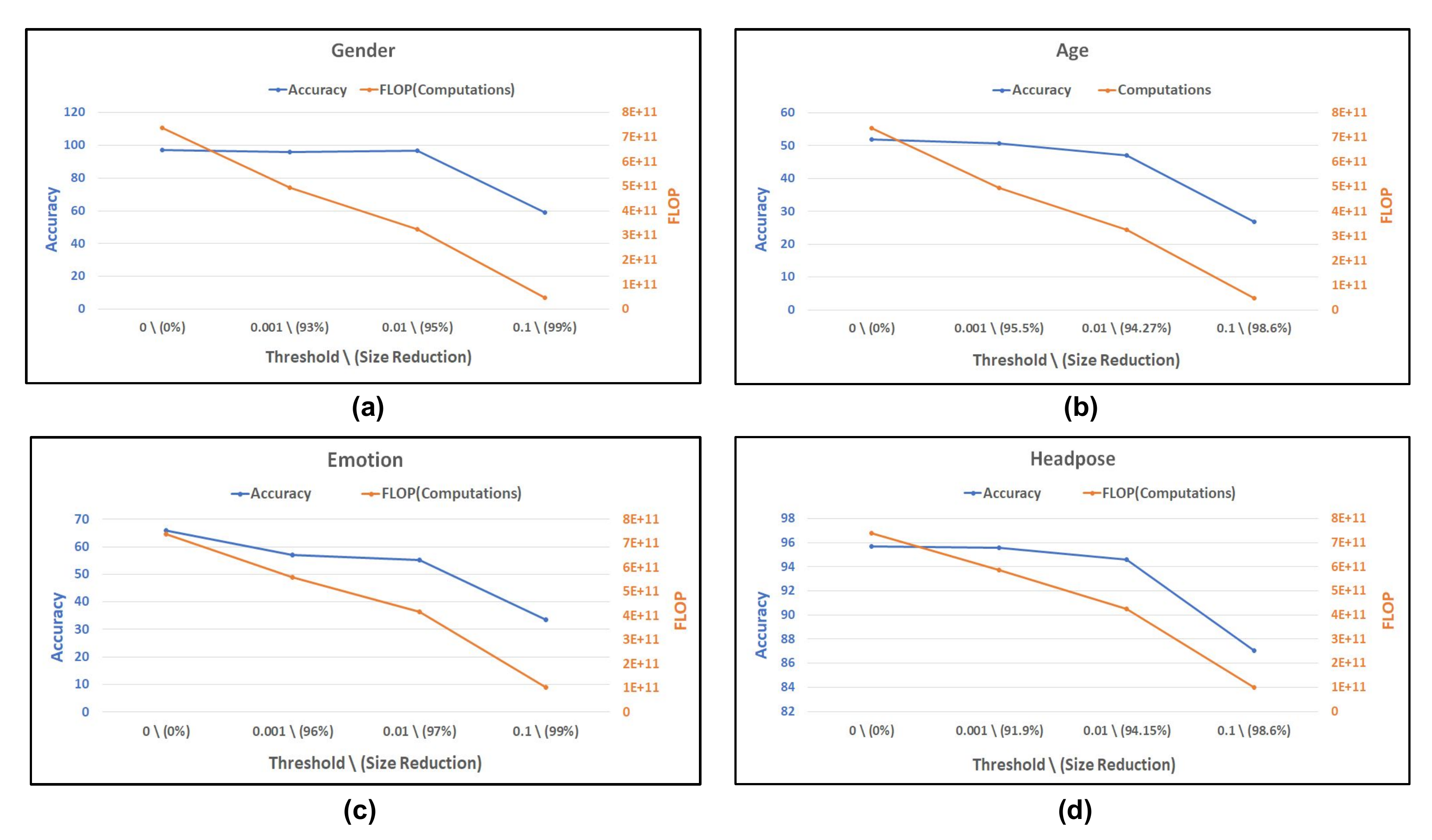}
    \caption{The four figures show the accuracy and computational complexity for VGG-Face network pruned with different thresholds. For each task, the threshold was varied from 0.1 to 0.001. A threshold of 0 indicates the finetuned network which is not pruned. We have shown the accuracy on the left axis and computational cost (number of flops) on the right axis. The percentage reduction in size is given along with the respective threshold values on the X-axis. a) Gender b) Age c) Emotion d) Head pose}
    \label{fig:prunefinetune}
\end{figure*}

\begin{figure*}
    \centering
    \includegraphics[width = \linewidth]{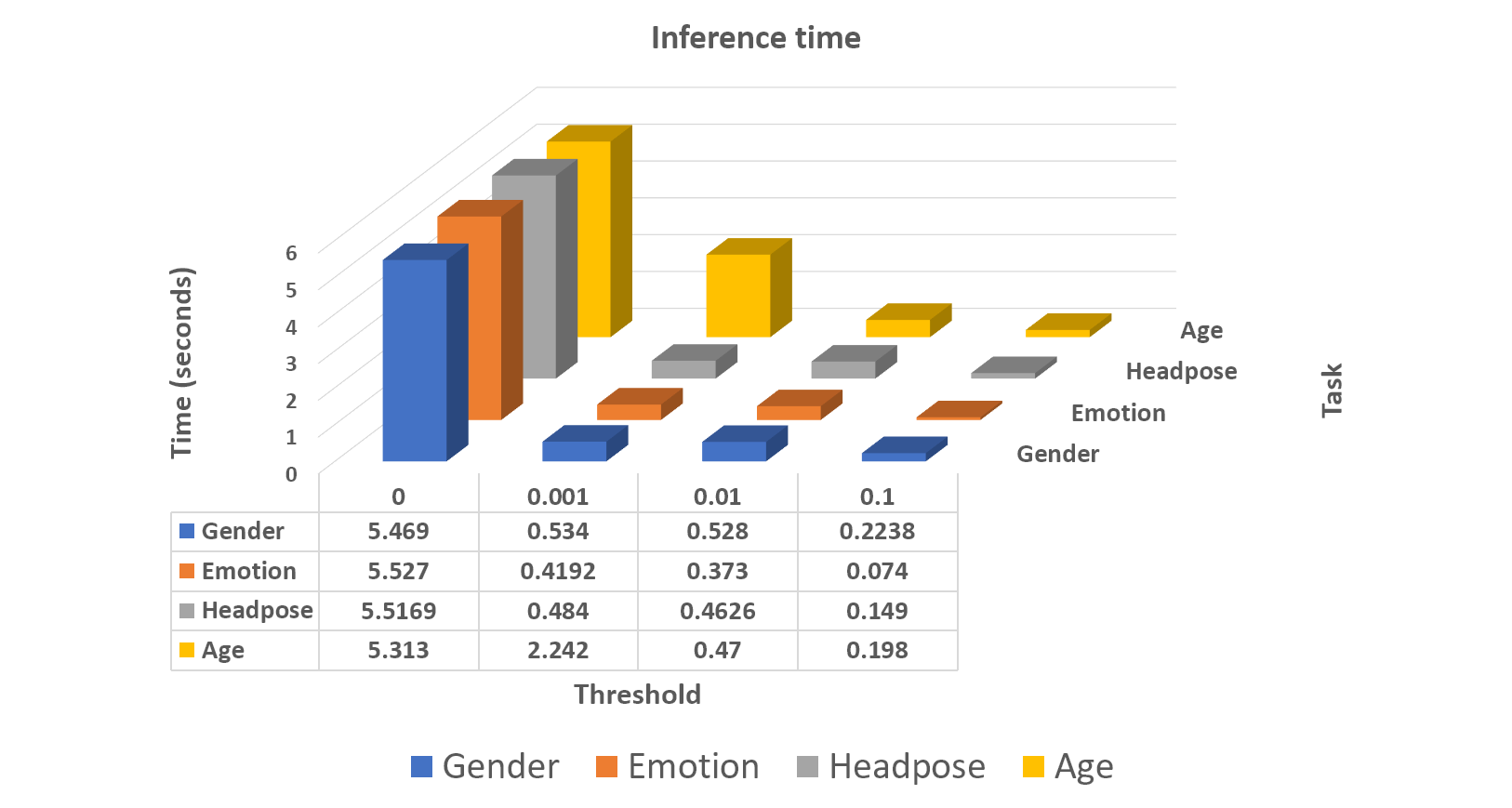}
    \caption{Time for inference of one image on the CPU for the full network versus pruned networks.}
    \label{fig:inferencetime}
\end{figure*}

We also show results of finetuning the pruned networks on the data sets of the satellite tasks. This allows us to get an accuracy similar to the uncompressed network. The networks were pruned with threshold values of 0.1,  0.01 and 0.001 (See Equation \ref{eqn:kneepoint}). The results are shown in Figure \ref{fig:prunefinetune}. We observe that as threshold increases, the network size and computational complexity reduces significantly while retaining the accuracy. Thus, threshold is a reliable way to tune the pruning algorithm and get the desired compromise between compression ratio and accuracy. We also note that the best threshold for all tasks is 0.01, as it gives a good trade-off between accuracy and compression ratio. The inference time of compressed networks is 10 times faster than the original networks, while giving similar accuracy, as seen in Figure \ref{fig:inferencetime}. 

\section{Discussion}
\begin{figure*}
    \centering
    \includegraphics[width = \linewidth]{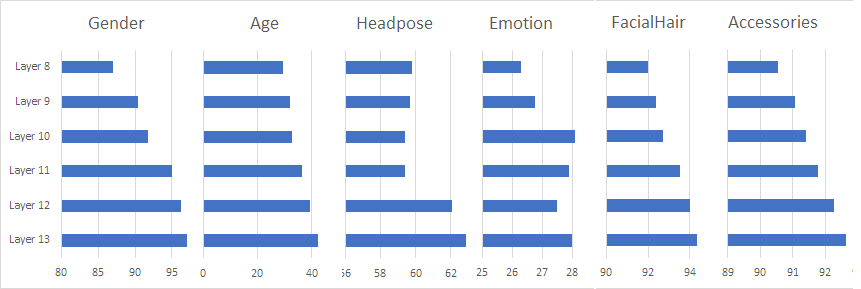}
   \caption{Figure shows the accuracy obtained by regressing the activations of the last 6 convolutional layers of face recognition network, VGG-Face, on various satellite tasks. }
    \label{fig:vgglayers}
\end{figure*}

\paragraph{Key Observations:} 
Some of our major observations can be summarized as follows:
\begin{itemize}
\item We can adapt the features learned by a deep network for a  task it was not trained for, without having access to the original data set.  The complex features learned on  a large data set can be leveraged  for tasks for which large data sets are not available. This works well in practice.
\item Most of the face tasks are highly related to each other. Thus we can easily transfer knowledge among them. Models learned on certain tasks such as face recognition are very versatile and give good accuracy when transferred to other tasks. Emotion detection models are not very useful for transferring to other face tasks (and seem to be largely on their own in our list of face tasks).
\item The information pertaining to other tasks is encoded by very few filters in each layer of a network. thus in most cases, we achieve very high level of compression by removing redundant filters.
\item This pattern of having few useful filters is present in other architectures such as LightCNN \cite{lightcnn} and networks trained using other loss functions. Our approach corroborates these earlier efforts, and provides a simple methodology to adapt this insight for cross-task pruning.
\end{itemize}

\paragraph{Performance of Different Convolutional Layers:}

A network with VGG-16 architecture has 13 convolutional layers. We now try to ask: which layer gives the best results for transferring information to other tasks in our approach? Our regression experiments were conducted on all the convolutional layers of the network and best results were recorded.  The accuracy for the last 6 layers for different tasks is given in Figure \ref{fig:vgglayers}. The layer which gives the best result varies for different satellite tasks, as expected. For instance, head pose, gender and age are best learned from layer 13 while the task emotion is best learned from layer 10. The layer in which task is learned with best accuracy signifies its relation with the primary task.


\paragraph{Extension to Different Data Sets, Tasks and Architectures:}
Our framework can easily be extended to data sets and tasks other than what is explored in this paper. It can also be applied to any pre-trained network, even if we do not have the original data set it was trained for. In Section \ref{sec:pruning}, we use a pretrained VGG-Face network \cite{VGGFace} as our base network. All the satellite tasks were on varied data sets, showing that our framework can be extended to other CNN architectures, by finding redundant filters and removing them. 

\paragraph{Applications and Future Directions:}
The knowledge of the various tasks encoded by different filters opens up a lot of opportunities. This can be very useful for finetuning and transfer learning. Before training the network, determine which filters are needed and remove the rest of the filters. This reduces the resources and time needed for training. We can also use this knowledge to customize networks in creative ways. For example, if we want to train a network for emotion and render it agnostic to face identity, we can find which filters represent identity and remove those filters to get an identity-agnostic network.

In total, we trained 180 networks which took approximately 760 GPU hours on Nvidia GeForce GTX 1080 Ti. In addition, we performed 1690 linear regression experiments on CPU. All codes were implemented in PyTorch Deep Learning Framework.

\section{Conclusion}

In this work, we explored several tasks in the face domain and their relationship to each other. Our proposed methodology, which uses a humble linear regression model, allowed us to leverage networks trained on large data sets, such as face recognition networks, for satellite tasks that have less data, such as determining the age of a person, head pose, emotions, facial hair, accessories, etc. Our method provided a computationally simple method to adapt a pretrained network for a task it was not trained for. We were able to estimate where the information was stored in the network and how well the features transferred from the primary task to the satellite task. These insights were used to prune networks for a specific task. Our results showed that it is possible to achieve high compression rates with a slight reduction of performance.

{\small
\bibliographystyle{ieee}
\bibliography{egbib}
}

\end{document}